\documentclass{article}
\usepackage[utf8]{inputenc}
\usepackage[T1]{fontenc}
\usepackage{arxiv}
\usepackage{natbib}
\setcitestyle{authoryear,round,citesep={;},aysep={,},yysep={;}}
\usepackage{microtype}
\usepackage{float}

\usepackage{amsmath,amsfonts,bm}

\def\eqref#1{equation~\ref{#1}}

\def\1{\bm{1}}

\DeclareMathAlphabet{\mathsfit}{\encodingdefault}{\sfdefault}{m}{sl}
\SetMathAlphabet{\mathsfit}{bold}{\encodingdefault}{\sfdefault}{bx}{n}

\usepackage{graphicx}
\usepackage{booktabs}
\usepackage{multirow}
\usepackage{pgfplots}
\pgfplotsset{compat=1.18}
\definecolor{queryBlue}{HTML}{0072B2}
\definecolor{queryRed}{HTML}{D55E00}
\pgfplotsset{query sensitivity/.style={
    width=0.73\linewidth, height=2.5cm, scale only axis,
    xmin=0.5, xmax=12.5, xtick={1,3,5,7,9,12},
    xlabel={Mining queries ($K$)},
    tick label style={font=\fontsize{8}{9}\selectfont},
    label style={font=\fontsize{8}{9}\selectfont}, tick align=outside,
    scaled y ticks=false,
    yticklabel style={/pgf/number format/fixed},
    legend style={font=\fontsize{8}{9}\selectfont, draw=none, fill=none,
                  cells={anchor=west}, at={(0.5,1.05)}, anchor=south,
                  legend columns=-1, /tikz/every even column/.append style={column sep=2pt}},
    every axis plot/.append style={line width=0.9pt, mark size=1.7pt},
}}
\usepackage{amssymb}
\usepackage{hyperref}
\usepackage{url}

\title{FM-ReID: Selective Competitive Token Routing for Object Re-Identification}

\author{
    Zhiqi Li$^{1,2}$,
    Xiaowei Zhou$^{1}$\thanks{Corresponding authors: \texttt{zhouxiaowei@ouc.edu.cn} and \texttt{dongjunyu@ouc.edu.cn}.},
    Zeyuan Sun$^{1}$,
    Feng Gao$^{1}$,
    Junyu Dong$^{1,2}$\footnotemark[1] \\[0.5em]
    $^{1}$Faculty of Information Science and Engineering, 
    Ocean University of China, Qingdao, China \\
    $^{2}$Sanya Oceanographic Institution, 
    Ocean University of China, Sanya, China \\[0.5em]
    \texttt{\{lizhiqi,sunzeyuan\}@stu.ouc.edu.cn} \\
    \texttt{\{zhouxiaowei,gaofeng,dongjunyu\}@ouc.edu.cn}
}
\date{}

\renewcommand{\shorttitle}{FM-ReID: Selective Competitive Token Routing}
\hypersetup{
    pdftitle={FM-ReID: Selective Competitive Token Routing for Object Re-Identification},
    pdfauthor={Zhiqi Li, Xiaowei Zhou, Zeyuan Sun, Feng Gao, Junyu Dong},
    pdfsubject={Object re-identification},
    pdfkeywords={Object re-identification, Fine-grained representation learning, Competitive token routing, Visual foundation models, Metric learning}
}
\begin{document}

\maketitle

\begin{abstract}

Object re-identification (ReID) faces a recurring challenge: different identities can share highly similar global appearances, while the cues that distinguish them are localized, heterogeneous, and visible only under particular viewpoints. This challenge arises in animal ReID through markings, contours, and scars, in person ReID through subtle clothing and accessory cues, and in vehicle ReID through localized appearance details. Although visual foundation models encode such information in dense tokens, a single holistic descriptor can obscure discriminative local signals. We propose FM-ReID, an end-to-end framework that formulates local representation learning as selective competitive token routing. Its Competitive Fine-grained Mining module uses multiple mining queries and a residual query to compete for dense DINOv3 tokens. Above-prior selection retains tokens preferentially allocated to each mining query, while the residual slot receives tokens excluded from the retrieval descriptors. The resulting multi-query descriptors are jointly trained with a holistic representation for retrieval, without fixed spatial partitions or equal-area constraints. FM-ReID achieves strong results on animal, person, and vehicle ReID benchmarks, supporting competitive token routing as an effective way to augment holistic foundation-model representations.
\end{abstract}

\begin{figure}[H]
    \centering
    \includegraphics[width=1\linewidth]{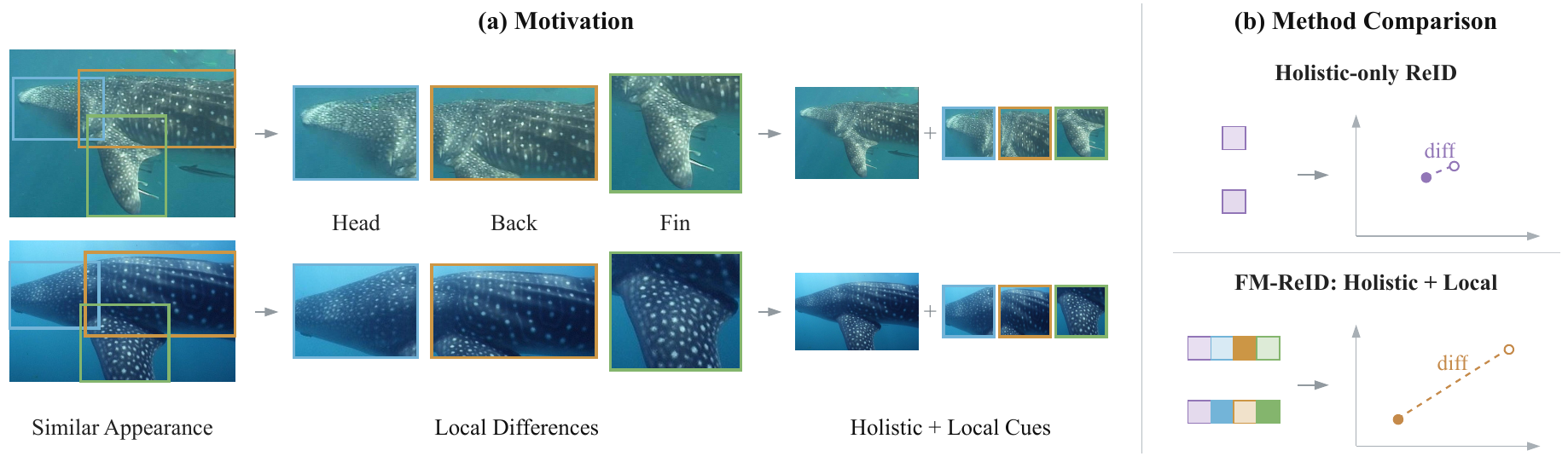}
    \caption{(a) Object ReID often requires localized identity cues to complement holistic representations when different instances have similar global appearances.  (b) FM-ReID selectively mines such cues from dense foundation-model tokens through competitive query routing, without fixed spatial partitions or equal-area constraints.}
    \label{fig:motivation}
\end{figure}

\section{Introduction}

Object re-identification (ReID) retrieves the same instance across cameras, times, or viewpoints. Its central difficulty is that different identities can share highly similar global appearances, whereas the evidence that distinguishes them is often localized, heterogeneous, and visible only from particular viewpoints. In animal ReID, this evidence may consist of markings, contours, or scars~\citep{wildlifedatasets,wildlifereid10k}; in person ReID, it can be subtle clothing and accessory cues; and in vehicle ReID, it can be localized appearance details. Viewpoint changes, deformation, occlusion, and background clutter further alter the location and visibility of these cues.

Local and multi-granular representation learning has long been used to complement holistic features with additional localized representations. Early approaches divide feature maps into fixed stripes or multiple predetermined granularities~\citep{pcb,mgn}, while more recent Transformer-based methods rearrange patch tokens or learn prototypes that attend to latent regions~\citep{transreid,pat,aaformer}. However, fixed partitions and rearrangement rules rely on approximate spatial alignment and can become unreliable across shapes, poses, and occlusions. Independently normalized queries may also repeatedly attend to the same salient evidence. These limitations motivate a local representation mechanism that can select complementary evidence without fixed spatial partitions, equal-area constraints, or category-specific anatomy.

Self-supervised visual foundation models provide a promising basis for this mechanism. DINO-family models learn object-aware dense representations from large-scale data without category-specific part annotations~\citep{dino,dinov2,dinov3}. Although their patch tokens encode rich spatial and semantic structure, dense tokens alone do not determine which local evidence should augment a retrieval descriptor. The allocation mechanism must distinguish query-specific evidence from ambiguous or uninformative tokens, while accommodating identity cues with unequal visible extents.

We address this problem with FM-ReID, an end-to-end framework that formulates local representation learning as selective competitive token routing on top of a DINOv3 visual encoder. Its holistic branch fuses the image-level CLS token with average-pooled patch context. Its Competitive Fine-grained Mining (CFM) module introduces multiple learnable mining queries and a residual query that compete for dense patch tokens through query-wise normalization. Above-prior selection retains tokens preferentially allocated to each mining query before query-specific aggregation, whereas the residual slot receives evidence excluded from the retrieval descriptors. Thus, a token need not contribute to any mining descriptor, and the retained support can vary across queries. We jointly train the resulting multi-query descriptors with the holistic representation for retrieval, yielding a multi-granular embedding without fixed spatial partitions or equal-area constraints.

Our main contributions are summarized as follows:
\begin{itemize}
    \item We propose FM-ReID, an end-to-end object ReID framework that augments a holistic representation with fine-grained features selectively mined from dense foundation-model tokens, without part annotations.
    \item We introduce the CFM module, in which mining queries and a residual query compete for tokens; above-prior selection allows evidence to remain outside the retrieval descriptors and permits query supports of unequal size.
    \item We jointly train holistic and multi-query descriptors for retrieval. Evaluations on animal, person, and vehicle ReID benchmarks show that competitive token routing can effectively augment holistic foundation-model representations.
    \item We provide FM-FISH, a benchmark for fine-grained identity discrimination among visually similar fish in underwater and laboratory settings.
\end{itemize}

\section{Related Work}

\subsection{Fine-Grained Representation Learning for Object ReID}

Fine-grained representations are crucial when identities share similar holistic appearances. The challenge is especially pronounced in animal ReID, where localized markings and subtle structural differences may be the only reliable identity cues~\citep{wildlifedatasets,wildlifereid10k}, but analogous ambiguities occur in person and vehicle ReID.

Early part-based methods impose spatial structure directly. PCB~\citep{pcb} partitions convolutional feature maps into horizontal stripes, while MGN~\citep{mgn} combines global features with local descriptors at several fixed granularities. Such partitions are effective for approximately aligned pedestrians, but can become semantically inconsistent under pose changes, occlusion, or non-human object geometry. Transformer-based methods provide more flexible token processing. TransReID~\citep{transreid} shifts, shuffles, and groups patch tokens through its Jigsaw Patch Module. PAT~\citep{pat} learns part prototypes with a Transformer decoder, and AAformer~\citep{aaformer} uses Optimal Transport to align patches with learnable part tokens. For animals, PAW-ViT~\citep{pawvit} specializes anatomical part tokens through semantic segmentation distillation. Our propovsed FM-ReID uses identity supervision to learn selective token aggregation without prescribed anatomical regions or equal-area assignments.

\subsection{Query-Based Discovery and Competitive Assignment}

Learnable queries and prototypes have been used to discover latent regions from weak supervision. Region Grouping~\citep{regiongrouping} learns interpretable object regions from image-level labels, while conceptual-part models~\citep{conceptparts} associate spatial features with latent prototypes. In ReID, PAT applies part prototypes independently and introduces auxiliary objectives to improve diversity. APD~\citep{apd} uses competition in part-mask generation, and AAformer adopts a globally balanced patch-to-part assignment. DRL-Net~\citep{drlnet} separates identity-relevant and identity-irrelevant queries. These methods establish prior uses of competitive assignment and auxiliary queries; CFM combines them with above-prior selection over dense foundation-model tokens.

Competition among latent slots has also been studied in object-centric learning. Slot Attention~\citep{slotattention} normalizes attention across slots and renormalizes over inputs for aggregation. Our proposed CFM builds on this allocation principle for discriminative retrieval. Its residual query and above-prior selection allow evidence to remain unclaimed by the mining queries, while query-specific aggregation permits unequal support sizes. The intended contribution is this selective aggregation of additional identity evidence from strong dense representations.

\subsection{Visual Foundation Models for ReID}

Vision-language models have become strong initialization sources for ReID. CLIP-ReID~\citep{clipreid} learns identity-specific text tokens and transfers textual knowledge to image representations. TF-CLIP~\citep{tfclip} constructs visual memories for video ReID, CLIMB-ReID~\citep{climbreid} combines CLIP-based transfer with sequence modeling, and EvoPrompt-ReID~\citep{evopromptreid} jointly evolves prompts and the image encoder through bilevel optimization. Their main emphasis is cross-modal knowledge or temporal modeling rather than competitive organization of dense spatial tokens.

A complementary direction learns person-specific representations through additional self-supervised pre-training. LUPerson~\citep{luperson} supplies large-scale unlabeled pedestrian data; PASS~\citep{pass} learns part-aware consistency from predefined local zones; and SOLIDER~\citep{solider} introduces pseudo-semantic supervision and controllable representation learning. Although effective, these approaches use in-domain data, person-oriented spatial assumptions, or additional pre-training stages.

General self-supervised foundation models offer category-agnostic dense features. DINO~\citep{dino} reveals emergent object structure in self-supervised Vision Transformers, DINOv2~\citep{dinov2} scales the resulting general-purpose representations, and register tokens improve the spatial consistency of dense feature maps~\citep{dinoregister}. DINOv3~\citep{dinov3} further advances dense self-supervised features at scale. Our proposed FM-ReID uses this visual knowledge without text prompts or secondary in-domain pre-training. It augments holistic pooling with selective competitive aggregation, whose empirical contribution is evaluated using the same DINOv3 backbone.

\begin{figure}[t]
    \centering
    \includegraphics[width=1\linewidth]{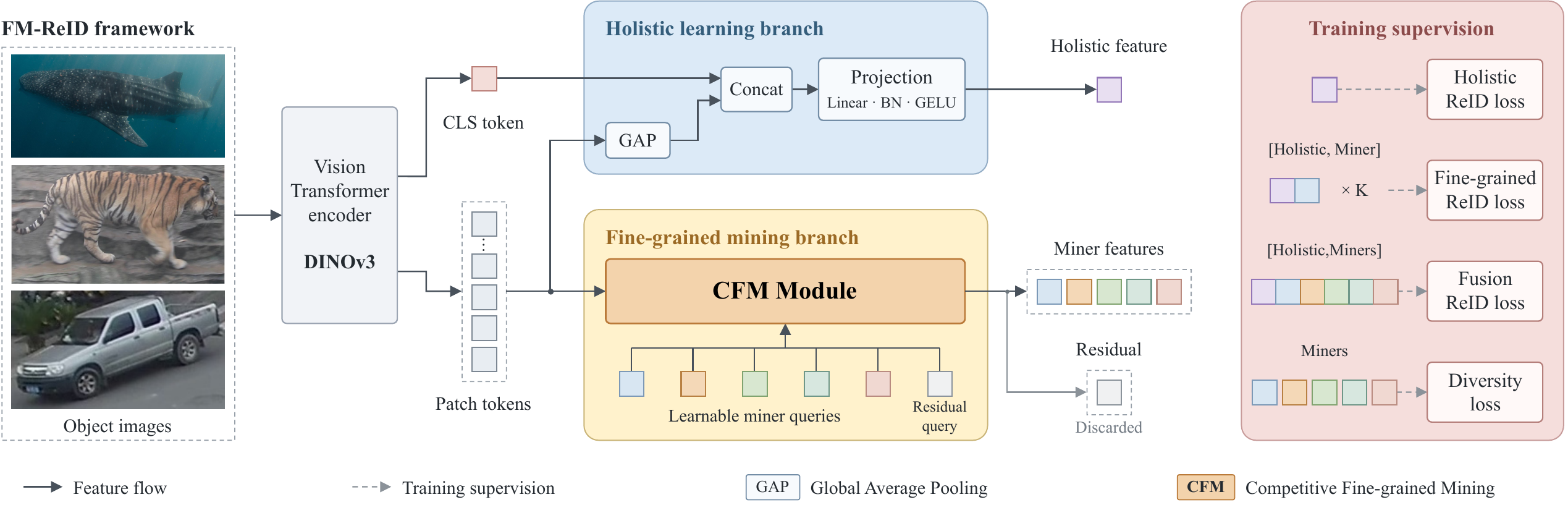}
    \caption{Overview of FM-ReID. DINOv3 provides a holistic CLS token and dense patch tokens. CFM selectively routes patch-token evidence through competing mining and residual queries, then retains above-prior assignments for the mining descriptors. During training, the holistic feature, holistic--fine-grained pairs, and their joint fusion receive ReID supervision. At inference, normalized holistic and fine-grained features form the retrieval embedding, while the residual output is discarded.}
    \label{fig:framework}
\end{figure}

\section{Method}
\label{sec:method}

\subsection{Overview of FM-ReID}

FM-ReID augments a holistic representation with selectively mined localized representations. As illustrated in Figure~\ref{fig:framework}, an input image $\mathbf{X}\in\mathbb{R}^{H\times W\times C}$ is processed by a DINOv3 encoder, which produces a CLS token $\mathbf{z}_{\mathrm{cls}}\in\mathbb{R}^{D}$ and $N$ spatial patch tokens $\mathbf{Z}=[\mathbf{z}_1,\ldots,\mathbf{z}_N]\in\mathbb{R}^{N\times D}$. We discard register tokens because they do not correspond to image locations. The holistic branch combines the CLS token with average-pooled patch context:
\begin{equation}
\mathbf{f}_{h}=\phi_h\!\left(
 [\mathbf{z}_{\mathrm{cls}};N^{-1}\textstyle\sum_{n=1}^{N}\mathbf{z}_n]
 \right),
\label{eq:holistic}
\end{equation}
where $[\cdot;\cdot]$ denotes concatenation and $\phi_h$ is a linear--BN--GELU projection. In parallel, the Competitive Fine-grained Mining (CFM) module converts $\mathbf{Z}$ into $K$ fine-grained representations. Rather than assigning queries to fixed spatial regions, CFM makes mining queries and a residual query compete for each patch token, then retains only above-prior mining assignments for aggregation. The residual output is not used as a retrieval descriptor, providing an alternative allocation for evidence that is not preferentially retained by the mining queries. The following two subsections detail this selective aggregation and the joint supervision used to combine the holistic and fine-grained representations for retrieval. All components, including the DINOv3 encoder, are optimized end to end.

\subsection{Competitive Fine-Grained Mining}

Independent attention queries can repeatedly select the same dominant region because each query normalizes its attention over all tokens in isolation. CFM instead makes its queries compete for every token before aggregation.

\paragraph{Competitive token assignment.}
We use $K$ learnable fine-grained mining queries $\{\mathbf{q}_k\}_{k=1}^{K}$ and one residual competition query $\mathbf{q}_0$. In the reported experiments, the mining-query matrix is initialized as $\mathbf{G}/\|\mathbf{G}\|_F$, where the entries of $\mathbf{G}\in\mathbb{R}^{K\times D}$ are sampled from a standard Gaussian. The residual query is independently initialized from $\mathcal{N}(0,0.02^2\mathbf{I}_D)$ and provides an auxiliary non-retrieval slot for competing assignments. Let $\mathbf{Q}=[\mathbf{q}_0;\mathbf{q}_1;\ldots;\mathbf{q}_K]$. For attention head $m$, queries, keys, and values are computed as
\begin{equation}
\begin{array}{l}
\mathbf{Q}^{m}=\mathbf{Q}\mathbf{W}_{q}^{m},\quad
\mathbf{K}^{m}=(\mathbf{Z}+\mathbf{E})\mathbf{W}_{k}^{m},\\
\mathbf{V}^{m}=\mathbf{Z}\mathbf{W}_{v}^{m},
\end{array}
\label{eq:qkv}
\end{equation}
where $\mathbf{E}$ is a learnable spatial position embedding. Position is injected only into the keys, so it guides assignment without altering the visual content being aggregated.

Given the scaled similarities $s_{kn}^{m}=\langle\mathbf{Q}_{k}^{m},\mathbf{K}_{n}^{m}\rangle/\sqrt{d}$, where $d$ is the head dimension, CFM applies softmax across queries rather than tokens:
\begin{equation}
a_{kn}^{m}=\frac{\exp(s_{kn}^{m})}
{\sum_{j=0}^{K}\exp(s_{jn}^{m})},\qquad
\sum_{k=0}^{K}a_{kn}^{m}=1.
\label{eq:competition}
\end{equation}
Consequently, increasing one query's assignment to a patch necessarily reduces the mass available to the others. The residual output is excluded from direct identity supervision and the retrieval descriptor. The residual query still receives gradients indirectly through the shared competition.

\begin{figure}[t]
    \centering
    \includegraphics[width=1\linewidth]{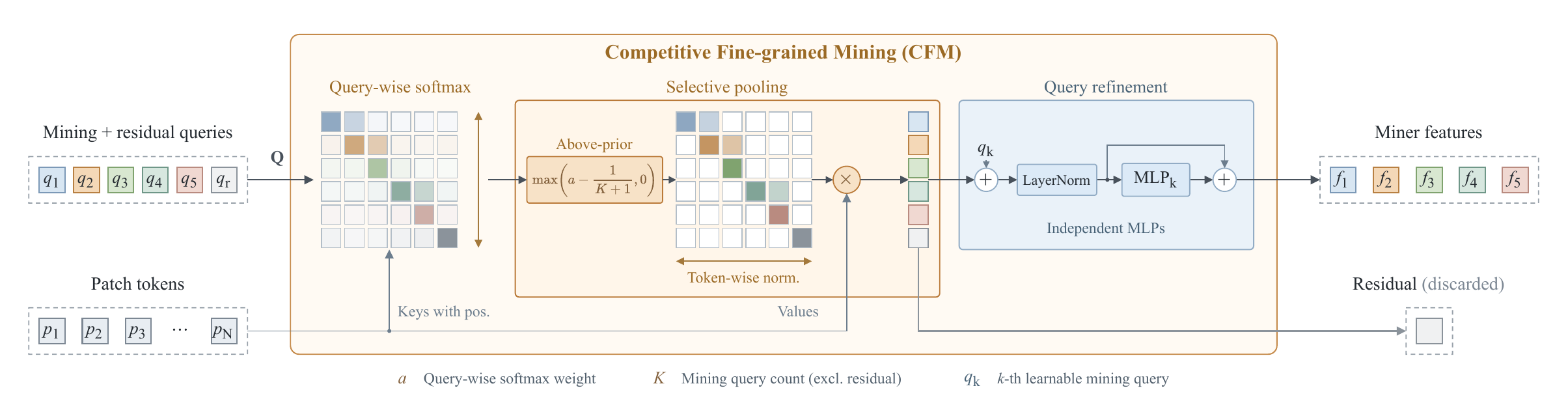}
    \caption{Competitive Fine-grained Mining (CFM). Mining queries and a residual query compete for each patch token through a softmax across queries. Above-prior mining assignments are retained and renormalized over tokens to form query-specific descriptors; their supports may overlap and vary in size. Query refinement produces the fine-grained features, whereas the residual output is excluded from the retrieval embedding.}
    \label{fig:cfm}
\end{figure}

\paragraph{Evidence selection and aggregation.}
A query-wise softmax may still produce ambiguous assignments close to the uniform solution. We therefore subtract the uniform competition prior, suppress assignments that do not exceed it, and renormalize each query over tokens:
\begin{equation}
\begin{array}{l}
\hat a_{kn}^{m}=\left[a_{kn}^{m}-\frac{1}{K+1}\right]_{+},\\
w_{kn}^{m}=\displaystyle\frac{\hat a_{kn}^{m}}
{\sum_{t=1}^{N}\hat a_{kt}^{m}+\epsilon}.
\end{array}
\label{eq:aggregation_weights}
\end{equation}
The first normalization in Equation~\ref{eq:competition} allocates probability mass across queries; the second in Equation~\ref{eq:aggregation_weights} pools each query's retained evidence. Multiple queries may exceed the prior for the same token, so their supports can overlap and need not have equal sizes. A token with no above-prior mining assignment contributes to none of the fine-grained aggregates. The multi-head output is
\begin{equation}
\mathbf{o}_k=\mathbf{W}_o\mathop{\mathrm{Concat}}_{m}
\left(\sum_{n=1}^{N}w_{kn}^{m}\mathbf{V}_{n}^{m}\right).
\label{eq:aggregation}
\end{equation}
After discarding $\mathbf{o}_0$, each fine-grained output is refined with an independent two-layer MLP:
\begin{equation}
\begin{array}{l}
\tilde{\mathbf{f}}_k=\mathrm{LN}(\mathbf{o}_k+\mathbf{q}_k),\\
\mathbf{f}_k=\tilde{\mathbf{f}}_k+\mathrm{MLP}_k(\tilde{\mathbf{f}}_k).
\end{array}
\label{eq:refinement}
\end{equation}
The residual connection adds the learned query seed, and query-specific refiners transform the aggregated features before ReID supervision.

\subsection{Optimization and Inference}

Let $\mathcal{R}(\mathbf{x},y)=\mathcal{L}_{\mathrm{id}}(\mathrm{BN}(\mathbf{x}),y)+\mathcal{L}_{\mathrm{tri}}(\mathbf{x},y)$ denote the standard ReID objective, comprising label-smoothed identity classification and batch-hard triplet loss. Separate BNNecks and classifiers are used for different feature streams. We supervise the holistic feature directly. For each mined feature, we concatenate its holistic context $\mathbf{g}_k=[\mathbf{f}_h;\mathbf{f}_k]$ to stabilize identity learning. We also supervise their joint fusion $\mathbf{f}_u=[\mathbf{f}_h;\alpha\mathbf{f}_1;\ldots;\alpha\mathbf{f}_K]$, where $\alpha$ prevents the fine-grained blocks from overwhelming the holistic feature. The three ReID losses are
\begin{equation}
\begin{array}{l}
\mathcal{L}_{h}=\mathcal{R}(\mathbf{f}_h,y),\\
\mathcal{L}_{f}=\displaystyle\frac{1}{K}\sum_{k=1}^{K}\mathcal{R}(\mathbf{g}_k,y),\\
\mathcal{L}_{u}=\mathcal{R}(\mathbf{f}_u,y).
\end{array}
\label{eq:reid_objectives}
\end{equation}

To encourage decorrelation among mined feature vectors, let $\bar{\mathbf{F}}_f^{(b)}\in\mathbb{R}^{K\times D}$ stack the $L_2$-normalized fine-grained features of sample $b$. We regularize their Gram matrix by
\begin{equation}
\mathcal{L}_{\mathrm{div}}=\frac{1}{BK^2}\sum_{b=1}^{B}
\left\|\bar{\mathbf{F}}_f^{(b)}\bar{\mathbf{F}}_f^{(b)\top}
-\mathbf{I}_K\right\|_F^2.
\label{eq:diversity}
\end{equation}
The final training objective is
\begin{equation}
\mathcal{L}=\mathcal{L}_{h}+\lambda_u\mathcal{L}_{u}
+\lambda_f\mathcal{L}_{f}+\lambda_d\mathcal{L}_{\mathrm{div}}.
\label{eq:total_loss}
\end{equation}

At inference, the holistic and fine-grained BNNeck outputs are individually $L_2$-normalized and concatenated:
\begin{equation}
\mathbf{f}_{\mathrm{test}}=[\bar{\mathbf{f}}_h;
\alpha\bar{\mathbf{f}}_1;\ldots;\alpha\bar{\mathbf{f}}_K].
\label{eq:test_feature}
\end{equation}
The stored descriptor has $(K+1)D$ dimensions: 4,608 for the default $K=5$ and $D=768$, compared with 768 for the holistic baseline. Competition maps and the residual output are not stored for retrieval.

\section{Experiments}

\subsection{Experimental Setup}

\begin{table*}[t]
\centering
\fontsize{8.5}{10}\selectfont
\caption{Comparison on (a) WildlifeReID-10k under the official closed-set protocol and (b) vehicle ReID benchmarks. The first four baselines in (a) are from~\citet{wildlifereid10k}; TransReID and CLIP-ReID are from our runs. Bold indicates the best result in each metric.}
\label{tab:compact_sota}
\label{tab:vehicle_sota}
\begin{minipage}[t]{0.47\linewidth}
\vspace{0pt}
\centering
\textbf{(a) WildlifeReID-10k}\par\smallskip
\fontsize{8.5}{10}\selectfont
\setlength{\tabcolsep}{2pt}
\begin{tabular*}{\linewidth}{@{\extracolsep{\fill}}l c c c c@{}}
\toprule
Model & Input & mTop-1 & mTop-5 & BAKS \\
\midrule
ConvNeXt-Base & $224^2$ & 81.1 & 90.3 & 78.5 \\
EfficientNet-B3 & $300^2$ & 77.8 & 88.5 & 75.1 \\
ViT-Base & $224^2$ & 78.3 & 88.7 & 75.8 \\
Swin-Base & $224^2$ & 81.5 & 90.4 & 79.1 \\
\midrule
TransReID & $256^2$ & 80.52 & 89.46 & 78.22 \\
CLIP-ReID & $256^2$ & 76.49 & 87.61 & 74.21 \\
\textbf{FM-ReID} & $256^2$ & \textbf{85.03} & \textbf{92.07} & \textbf{83.33} \\
\bottomrule
\end{tabular*}
\end{minipage}\hfill
\begin{minipage}[t]{0.51\linewidth}
\vspace{0pt}
\centering
\textbf{(b) Vehicle ReID}\par\smallskip
\fontsize{8.5}{10}\selectfont
\setlength{\tabcolsep}{2pt}
\begin{tabular*}{\linewidth}{@{\extracolsep{\fill}}l c cc cc@{}}
\toprule
\multirow{2}{*}{Method} & \multirow{2}{*}{Reference} & \multicolumn{2}{c}{VeRi-776} & \multicolumn{2}{c}{VehicleID} \\
\cmidrule(lr){3-4}\cmidrule(lr){5-6}
& & mAP & R1 & R1 & R5 \\
\midrule
TransReID & ICCV21 & 82.1 & 97.4 & 83.6 & 97.1 \\
DCAL & CVPR22 & 80.2 & 96.9 & -- & -- \\
CLIP-ReID & AAAI23 & 83.3 & 97.4 & 85.3 & 97.6 \\
Fast-ReID & MM23 & 81.9 & 97.9 & 86.6 & 97.9 \\
EvoPrompt-ReID  & CVPR26 & 84.2 & 97.5 & - & - \\
\midrule
\textbf{FM-ReID} & -- & \textbf{87.8} & \textbf{98.3} & \textbf{87.4} & \textbf{98.6} \\
\bottomrule
\end{tabular*}
\end{minipage}
\end{table*}

\paragraph{Dataset}
We evaluate FM-ReID on animal, person, and vehicle ReID benchmarks. For animal ReID, we use WildlifeReID-10k~\citep{wildlifereid10k} and our FM-FISH benchmark. WildlifeReID-10k covers 37 source datasets. For person ReID, we use Market-1501~\citep{market1501}, MSMT17~\citep{msmt17}, DukeMTMC-reID~\citep{dukemtmc}, and Occluded-Duke~\citep{occ_duke}. The latter evaluates retrieval under partial occlusion. For vehicle ReID, we use VeRi-776~\citep{veri776} and VehicleID~\citep{vehicleid}.

Our proposed FM-FISH combines self-collected images, an FS-48 subset, AAUZebraFish, WhaleSharkID, and MFT25 track crops, comprising 45,632 images and 724 identity labels (Table~\ref{tab:fm_fish}(a)). It targets fine-grained matching among visually similar fish. We retain source-specific splits: the self-collected subset uses unseen test identities, the Wildlife subsets include known-identity queries, and MFT25 uses within-sequence temporal splits with track-based labels.

\paragraph{Implementation details.} We fine-tune DINOv3 ViT-Base/16~\citep{dinov3} on each benchmark with a batch size of 64, using five mining queries and one residual query. Input images are resized to $256\times256$ for animal and vehicle ReID, and $256\times128$ for person ReID. On WildlifeReID-10k, we train for 120 epochs with AdamW. Ablation variants share the same backbone and training recipe within each benchmark. All FM-ReID results use final checkpoints without re-ranking. All experiments are conducted on an NVIDIA GeForce RTX 5080 GPU.

\paragraph{Evaluation metrics.} We report mean average precision (mAP) and cumulative matching characteristic (CMC) accuracy for person, vehicle, and fish ReID following the respective benchmark protocols. For WildlifeReID-10k, we report mTop-1, mTop-5, and balanced accuracy on known samples (BAKS) under the official closed-set protocol~\citep{wildlifereid10k}. mTop-1 and mTop-5 are the unweighted means of the image-level top-1 and top-5 identification accuracies across source datasets, respectively. BAKS averages per-identity top-1 accuracy equally over known identities within each source dataset and then equally across source datasets.

\subsection{Evaluation on WildlifeReID-10k}

Table~\ref{tab:compact_sota}(a) compares FM-ReID with the published WildlifeReID-10k baselines and our TransReID and CLIP-ReID runs. FM-ReID achieves 85.03\% mTop-1, 92.07\% mTop-5, and 83.33\% BAKS, leading the listed methods on all three metrics. Compared with Swin-Base, the strongest official baseline, it improves mTop-1 and BAKS by 3.53 and 4.23 percentage points, respectively.

\subsection{Evaluation on Person ReID Benchmarks}
Table~\ref{tab:sota} compares FM-ReID with representative CNN- and Transformer-based methods on four person ReID benchmarks.

\begin{table*}[t]
\centering
\fontsize{8.5}{10}\selectfont
\caption{Comparison with selected published methods on MSMT17, Market-1501, DukeMTMC-reID, and Occ-Duke. \textbf{Bold} indicates the best result among the listed methods.}
\label{tab:sota}
\fontsize{8.5}{10}\selectfont
\setlength{\tabcolsep}{0pt}
\begin{tabular*}{\linewidth}{@{\extracolsep{\fill}} l l c cc cc cc cc}
\toprule
\multirow{2}{*}{Method} & \multirow{2}{*}{Reference} & \multirow{2}{*}{Backbone} & \multicolumn{2}{c}{MSMT17} & \multicolumn{2}{c}{Market-1501} & \multicolumn{2}{c}{DukeMTMC} & \multicolumn{2}{c}{Occ-Duke} \\
\cmidrule{4-5} \cmidrule{6-7} \cmidrule{8-9} \cmidrule{10-11} 
 & & & mAP & R1 & mAP & R1 & mAP & R1 & mAP & R1 \\
\midrule
PCB ~\cite{pcb} & ECCV18 & \multirow{3}{*}{CNN} & 40.4 & 68.2 & 81.6 & 93.8 & 69.2 & 83.3 & - & - \\
OSNet ~\cite{osnet} & ICCV19 &  & 52.9 & 78.7 & 84.9 & 94.8 & 73.5 & 88.6 & - & - \\
Fast-ReID ~\cite{fastreid} & MM23 &  & 59.9 & 83.3 & - & - & 78.9 & 89.6 & - & - \\
\midrule

TransReID ~\cite{transreid} & ICCV21 & \multirow{9}{*}{ViT} & 67.4 & 85.3 & 88.9 & 95.2 & 82.0 & 90.7 & 59.2 & 66.4 \\
DCAL ~\cite{dcal} & CVPR22 &  & 64.0 & 83.1 & 87.5 & 94.7 & 80.1 & 89.0 & - & - \\
DPM ~\cite{dpm} & MM22 &  & - & - & - & - & 82.6 & 91.0 & 61.8 & 71.4 \\
CLIP-ReID ~\cite{clipreid} & AAAI23 &  & 73.4 & 88.7 & 89.6 & 95.5 & 83.1 & 90.8 & 59.5 & 67.1 \\
TF-CLIP ~\cite{tfclip} & AAAI24 &  & 73.9 & 88.5 & 90.4 & 95.7 & - & - & - & - \\
AAformer ~\cite{aaformer} & TNNLS24 &  & 58.2 & 84.4 & 88.0 & 95.4 & 80.9 & 90.1 & - & - \\
EvoPrompt-ReID ~\cite{evopromptreid} & CVPR26 &  & 77.1 & 90.3 & 91.7 & 96.1 & - & - & - & - \\
\midrule
\textbf{FM-ReID (Ours)} & -- & ViT & \textbf{79.8} & \textbf{91.3} & \textbf{92.3} & \textbf{96.1} & \textbf{85.0} & \textbf{92.2} & \textbf{65.5} & \textbf{72.6} \\
\bottomrule
\end{tabular*}
\end{table*}

FM-ReID achieves the highest mAP among the listed methods on MSMT17, DukeMTMC-reID, and Occluded-Duke. On MSMT17, it obtains 79.8\% mAP and 91.3\% Rank-1, exceeding EvoPrompt-ReID by 2.7 and 1.0 percentage points, respectively. It also achieves competitive results of 92.3\% mAP and 96.1\% Rank-1 on Market-1501, and reaches 85.0\% mAP and 92.2\% Rank-1 on DukeMTMC-reID.

On Occluded-Duke, FM-ReID achieves 65.5\% mAP and 72.6\% Rank-1, improving over CLIP-ReID by 6.0 and 5.5 percentage points. These results demonstrate the framework's effectiveness in both standard and occluded person retrieval.

\subsection{Evaluation on Vehicle ReID Benchmarks}

As shown in Table~\ref{tab:vehicle_sota}(b), FM-ReID obtains 87.8\% mAP and 98.3\% Rank-1 on VeRi-776, improving mAP over EvoPrompt-ReID by 3.6 percentage points. On VehicleID, it reaches 87.4\% Rank-1 and 98.6\% Rank-5, exceeding Fast-ReID by 0.8 and 0.7 points, respectively. FM-ReID leads the listed methods on both vehicle benchmarks.

\subsection{Evaluation on FM-FISH}

\begin{table*}[t]
\centering
\fontsize{8.5}{10}\selectfont
\caption{FM-FISH: (a) dataset statistics and (b) comparison results. MFT25 identity labels denote annotated tracks. Bold indicates the best result in each metric.}
\label{tab:fm_fish}
\begin{minipage}[t]{0.43\linewidth}
\vspace{0pt}
\centering
\textbf{(a) FM-FISH dataset statistics}\par\smallskip
\fontsize{8.5}{10}\selectfont
\setlength{\tabcolsep}{2pt}
\begin{tabular*}{\linewidth}{@{\extracolsep{\fill}}lrr@{}}
\toprule
Source & IDs & Images \\
\midrule
Self-collected & 36 & 1,205 \\
FS-48 subset & 48 & 22,692 \\
AAUZebraFish & 5 & 281 \\
WhaleSharkID & 519 & 7,305 \\
MFT25 & 116 & 14,149 \\
\midrule
\textbf{Total} & \textbf{724} & \textbf{45,632} \\
\bottomrule
\end{tabular*}
\end{minipage}\hfill
\begin{minipage}[t]{0.55\linewidth}
\vspace{0pt}
\centering
\textbf{(b) ReID results}\par\smallskip
\fontsize{8.5}{10}\selectfont
\setlength{\tabcolsep}{2pt}
\begin{tabular*}{\linewidth}{@{\extracolsep{\fill}}l c cccc@{}}
\toprule
Method & Reference & mAP & R1 & R5 & R10 \\
\midrule
TransReID & ICCV21 & 43.2 & 53.7 & 65.8 & 71.8 \\
CLIP-ReID & AAAI23 & 41.4 & 55.2 & 67.5 & 72.8 \\
\midrule
\textbf{FM-ReID} & -- & \textbf{55.1} & \textbf{63.5} & \textbf{72.3} & \textbf{76.3} \\
\bottomrule
\end{tabular*}
\end{minipage}
\end{table*}

Table~\ref{tab:fm_fish}(b) reports the results on FM-FISH. FM-ReID achieves 55.1\% mAP and 63.5\% Rank-1, outperforming TransReID by 11.9 and 9.8 percentage points, and CLIP-ReID by 13.7 and 8.3 points, respectively. The gains across all reported metrics support its effectiveness for fine-grained fish identity matching.

\subsection{Ablation Studies}

\begin{table*}[t]
\centering
\fontsize{8.5}{10}\selectfont
\caption{Component ablation on WildlifeReID-10k and MSMT17. Mining and Residual denote the mining queries and auxiliary residual query. Within each benchmark, all variants use DINOv3-B/16, the same training recipe, and seed 1234; CFM variants retain the default fusion and diversity objectives. Bold indicates the best result in each metric.}
\label{tab:wildlife_ablation}
\fontsize{8.5}{10}\selectfont
\setlength{\tabcolsep}{2pt}
\begin{tabular*}{\linewidth}{@{\extracolsep{\fill}} l cc cc ccc cc}
\toprule
\multirow{2}{*}{Backbone} & \multicolumn{2}{c}{Holistic} & \multicolumn{2}{c}{CFM} & \multicolumn{3}{c}{WildlifeReID-10k} & \multicolumn{2}{c}{MSMT17} \\
\cmidrule(lr){2-3}\cmidrule(lr){4-5}\cmidrule(lr){6-8}\cmidrule(lr){9-10}
& CLS & GAP & Mining & Residual & mTop-1 & mTop-5 & BAKS & mAP & R1 \\
\midrule
\multirow{4}{*}{DINOv3-B/16}
& $\checkmark$ & & & & 83.29 & 91.44 & 81.46 & 76.1 & 89.6 \\
& $\checkmark$ & $\checkmark$ & & & 84.28 & 91.56 & 82.41 & 77.4 & 90.5 \\
\cmidrule(lr){2-10}
& $\checkmark$ & $\checkmark$ & $\checkmark$ & & 84.54 & 92.01 & 82.90 & 79.6 & \textbf{91.4} \\
& $\boldsymbol{\checkmark}$ & $\boldsymbol{\checkmark}$ & $\boldsymbol{\checkmark}$ & $\boldsymbol{\checkmark}$ & \textbf{85.03} & \textbf{92.07} & \textbf{83.33} & \textbf{79.7} & \textbf{91.4} \\
\bottomrule
\end{tabular*}
\end{table*}

Table~\ref{tab:wildlife_ablation} evaluates the contributions of holistic aggregation and fine-grained mining. Adding GAP to CLS improves WildlifeReID-10k mTop-1 by 0.99 points and MSMT17 mAP by 1.3 points. Full CFM further improves these metrics by 0.75 and 2.3 points over CLS+GAP. Retaining the residual query adds 0.49 points in WildlifeReID-10k mTop-1 and 0.1 points in MSMT17 mAP, supporting its role in competitive aggregation.

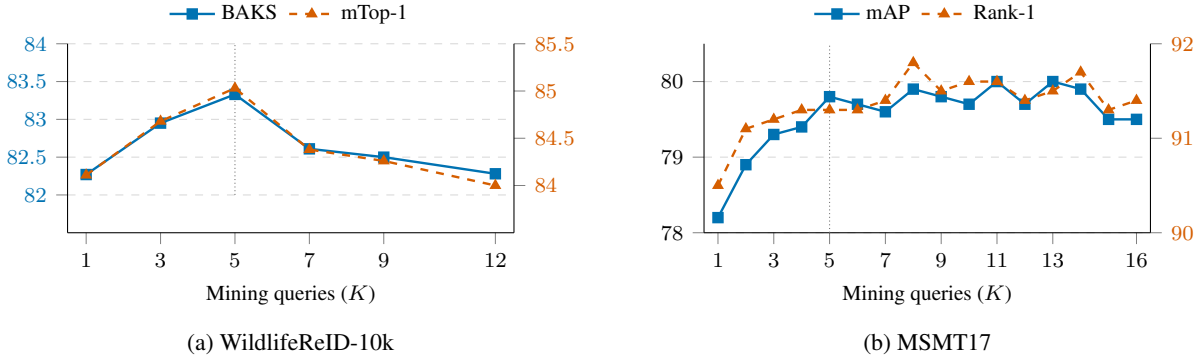
\begin{figure}[h]
\centering
\begin{minipage}[t]{0.49\linewidth}
\centering
\begin{tikzpicture}
\begin{axis}[query sensitivity,
    axis y line*=left, axis x line*=bottom,
    ylabel={},
    ymin=81.5, ymax=84, ytick={82,82.5,83,83.5,84},
    yticklabel style={color=queryBlue},
    ymajorgrids=true, grid style={dashed, black!15},
]
\draw[black!55, densely dotted] (axis cs:5,82) -- (axis cs:5,84);
\addplot[color=queryBlue, mark=square*, mark options={solid}]
    coordinates {(1,82.27) (3,82.95) (5,83.33) (7,82.61) (9,82.50) (12,82.28)};
\addlegendentry{BAKS}
\addlegendimage{color=queryRed, dashed, mark=triangle*, mark options={solid}}
\addlegendentry{mTop-1}
\end{axis}
\begin{axis}[query sensitivity,
    axis y line*=right, axis x line=none, xlabel={},
    ylabel={},
    ymin=83.5, ymax=85.5, ytick={84,84.5,85,85.5},
    yticklabel style={color=queryRed},
]
\addplot[color=queryRed, dashed, mark=triangle*, mark options={solid}]
    coordinates {(1,84.11) (3,84.68) (5,85.03) (7,84.38) (9,84.26) (12,84.00)};
\end{axis}
\end{tikzpicture}
\par\smallskip
{\small (a) WildlifeReID-10k}
\end{minipage}
\hfill
\begin{minipage}[t]{0.49\linewidth}
\centering
\begin{tikzpicture}
\begin{axis}[query sensitivity,
    axis y line*=left, axis x line*=bottom,
    xmin=0.5, xmax=16.5, xtick={1,3,5,7,9,11,13,16},
    ylabel={}, ymin=78, ymax=80.5, ytick={78,79,80},
    ymajorgrids=true, grid style={dashed, black!15},
]
\draw[black!55, densely dotted] (axis cs:5,78) -- (axis cs:5,80.5);
\addplot[color=queryBlue, mark=square*, mark options={solid}]
    coordinates {(1,78.2) (2,78.9) (3,79.3) (4,79.4) (5,79.8) (6,79.7) (7,79.6) (8,79.9) (9,79.8) (10,79.7) (11,80.0) (12,79.7) (13,80.0) (14,79.9) (15,79.5) (16,79.5)};
\addlegendentry{mAP}
\addlegendimage{color=queryRed, dashed, mark=triangle*, mark options={solid}}
\addlegendentry{Rank-1}
\end{axis}
\begin{axis}[query sensitivity,
    axis y line*=right, axis x line=none, xlabel={},
    xmin=0.5, xmax=16.5, xtick={1,3,5,7,9,11,13,16},
    ylabel={},
    ymin=90, ymax=92, ytick={90,91,92}, yticklabel style={color=queryRed},
]
\addplot[color=queryRed, dashed, mark=triangle*, mark options={solid}]
    coordinates {(1,90.5) (2,91.1) (3,91.2) (4,91.3) (5,91.3) (6,91.3) (7,91.4) (8,91.8) (9,91.5) (10,91.6) (11,91.6) (12,91.4) (13,91.5) (14,91.7) (15,91.3) (16,91.4)};
\end{axis}
\end{tikzpicture}
\par\smallskip
{\small (b) MSMT17}
\end{minipage}
\caption{Sensitivity to the number of mining queries $K$: (a) WildlifeReID-10k, with BAKS/mTop-1 on the left/right axes; (b) MSMT17, with mAP/Rank-1 on the left/right axes. Dotted lines mark $K=5$; the residual query is retained and excluded from $K$. All runs use epoch-120 checkpoints and seed 1234. Panel (a) retains the full-model result at $K=5$, while (b) uses independent sweep runs for every $K$.}
\label{fig:query_sensitivity}
\end{figure}

\paragraph{Number of mining queries.}
Figure~\ref{fig:query_sensitivity} evaluates the effect of $K$. On WildlifeReID-10k, performance improves from $K=1$ to $K=5$, where mTop-1 and BAKS reach 85.03\% and 83.33\%, respectively. On MSMT17, mAP rises from 78.2\% at $K=1$ to 79.8\% at $K=5$ and remains between 79.5\% and 80.0\% for larger $K$. The default $K=5$ is within 0.2 points of the best mAP and 0.5 points of the best Rank-1 result on MSMT17, providing a compact choice with competitive performance.

\subsection{Efficiency}

\begin{table*}[t]
\centering
\caption{Inference cost at $256\times128$, FP32, and batch size 1 on an RTX 5080. Baseline uses DINOv3 with CLS pooling; parentheses indicate increases over Baseline + GAP.}
\label{tab:efficiency}
\small
\setlength{\tabcolsep}{4pt}
\begin{tabular*}{\linewidth}{@{\extracolsep{\fill}}lrrr@{}}
\toprule
Method & Parameters (M) & GFLOPs & Peak memory (MiB) \\
\midrule
TransReID & 92.92 & 40.77 & 371.5 \\
CLIP-ReID & 86.14 & 22.78 & 346.1 \\
\midrule
Baseline & 85.66 & 23.40 & 341.5 \\
Baseline + GAP & 86.84 & 23.40 & 346.0 \\
\textbf{FM-ReID}
& {\scriptsize (+9.66\%)} 95.23 
& {\scriptsize (+1.41\%)} 23.73
& {\scriptsize (+9.28\%)} 378.1 \\
\bottomrule
\end{tabular*}
\end{table*}

Table~\ref{tab:efficiency} compares inference-model parameters, FLOPs, and peak GPU memory. Adding GAP to the CLS baseline introduces 1.18M parameters and 4.5 MiB of peak memory, with unchanged FLOPs at the reported precision. Relative to Baseline + GAP, FM-ReID increases FLOPs by only 1.41\%, with parameter and peak-memory increases of 9.66\% and 9.28\%, respectively. TransReID uses its checkpoint-specific patch stride of 12, while all other methods use 16; its higher FLOPs also reflect the longer token sequence.

\subsection{Qualitative Visualization}

\begin{figure}[h]
    \centering
    \includegraphics[width=\linewidth]{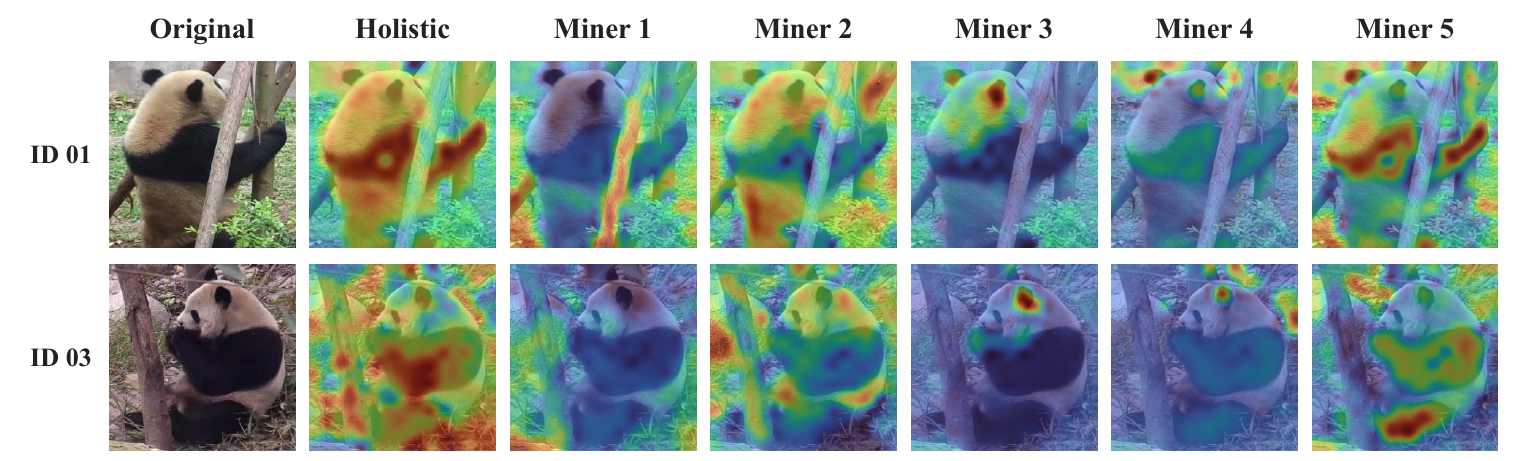}
    \caption{Qualitative visualization on two panda identities.
    From left to right: input images, holistic responses,
    and five miner attention maps. Holistic responses use
    feature--patch cosine similarity, while miner maps show
    token aggregation weights. Each map is independently
    min--max normalized for display; colors indicate relative
    responses within each map.}
    \label{fig:qualitative_visualization}
\end{figure}

Figure~\ref{fig:qualitative_visualization} shows the spatial
weighting patterns of the holistic representation and individual
miners. In these two examples, Miner~3 concentrates on the ear
region, while Miners~2 and~5 show more distributed responses over
the body and its surroundings. Miner~1 places greater emphasis
on surrounding structures. These examples illustrate
query-dependent aggregation of patch information from localized
regions, broader body areas, and contextual regions alongside
the holistic representation. We combine these additional miner features with the holistic
representation to enhance inter-identity separability in the
fused feature space and support more accurate re-identification.

\section{Conclusion}

We presented FM-ReID, an object ReID framework that extracts additional identity evidence from dense DINOv3 tokens through selective competition. Its CFM module combines mining queries, a residual query, and above-prior selection, allowing unclaimed evidence and unequal support sizes. The resulting features augment a holistic representation. FM-ReID exceeds the closed-set baselines reported by the WildlifeReID-10k benchmark paper, and within-backbone ablations evaluate its contribution beyond holistic pooling. Separate supervised experiments on animal, person, and vehicle benchmarks support the architecture's cross-category applicability. Transfer to unseen object categories remains unevaluated.

\bibliography{references}
\bibliographystyle{unsrtnat}

\end{document}